\documentclass{article}
\usepackage{spconf,amsmath,graphicx}
\usepackage{amsfonts}
\usepackage{hyperref}
\usepackage{bm}
\usepackage{float}
\usepackage{nicefrac}
\usepackage{support-caption}
\usepackage{subcaption}
\usepackage{multirow}
\usepackage{multicol}
\usepackage{makecell}
\usepackage{adjustbox}
\usepackage{booktabs}
\usepackage{tablefootnote}
\usepackage{xspace}

\title{Have Best of Both Worlds: \\ Two-Pass Hybrid and E2E Cascading Framework  for Speech Recognition}
\name{Guoli Ye, Vadim Mazalov, Jinyu Li {\rm and}  Yifan Gong}
 \address{Microsoft Corporation,  USA \\
  }
  
\begin{document}
\ninept
\maketitle
\begin{abstract}
Hybrid and end-to-end (E2E) systems have their individual advantages, with different error patterns in the speech recognition results. By jointly modeling audio and text, the E2E model performs better in matched scenarios and scales well with a large amount of paired audio-text training data. The modularized hybrid model is easier for customization, and better to make use of a massive amount of unpaired text data. This paper proposes a two-pass \emph{hybrid and E2E cascading (HEC)} framework to combine the  hybrid and E2E model in order to take advantage of both sides, with hybrid in the first pass and E2E in the second pass. We show that the proposed system achieves 8-10\% relative word error rate reduction with respect to each individual system. More importantly, compared with the pure E2E system, we show the proposed system has the potential to keep the advantages of hybrid system, e.g., customization and segmentation capabilities. We also show the second pass E2E model in HEC is robust with respect to the change in the first pass hybrid model.
\end{abstract}
\section{Introduction}
\label{sec:Introduction}
\vspace{-0.2cm}
End-to-end (E2E) modeling has gained significant success in automatic speech recognition (ASR) in recent years~\cite{soltau2016neural,li2018advancing, chiu2018state, he2019streaming, zhang2020transformer, li2020comparison,chen2021developing,gruenstein2021efficient}. Unlike conventional hybrid modeling that consists of several components developed independently, E2E modeling uses a single model to directly optimize the function of ASR. As a result, compared with hybrid models, E2E models can perform better when there is a large amount of paired audio and text training data, with the condition that the training and test scenarios match well \cite{li2021recent}.

Although E2E models benefit from the direct optimization, there are certain challenges in real life production scenarios. Hybrid modeling, thanks to its modularized design, can better address those challenges and offers some further advantages over E2E modeling. Firstly, given a personalized name or entity list, hybrid models are usually more effective in LM customization compared with E2E models.   Secondly, besides the paired audio and text training data, there usually exists a significantly larger amount of unpaired text-only data, which can easily be modeled by a language model (LM) in hybrid systems, but harder to be used in E2E models. Finally, the speech segmentation in hybrid systems is a very mature technique, while E2E models usually see accuracy loss by introducing end-of-speech detector. Despite a lot of efforts to tackle the issues above in E2E ASR systems~\cite{gruenstein2021efficient,toshniwal2018comparison,mcdermott2019density,meng2021internal,pundak2018deep,zhao2019shallow,wang2021light,li2021better,kim2021reducing}, it is fair to say that hybrid modelling is still more effective and practical to solve all the above problems due to
its modualized structure and technique maturity. As a result, as of today, hybrid models still dominate many commercial ASR applications.

Given that both hybrid and E2E modeling  have their own advantages, a natural question to ask is whether we can combine them to have the best of both worlds. Several publications explored this direction~\cite{li2019integrating,li2021combining,wong2020combination}. In these works, hybrid and E2E models are trained independently, and hypothesis level combination between the two systems is carried out by either rescoring the N-best and lattice from hybrid system with an E2E model~\cite{li2019integrating,li2021combining}, or Minimum Bayes' Risk (MBR) combination of the N-best lists of the two systems~\cite{wong2020combination}.

\begin{figure}[t]
    \centering
    \includegraphics[width=\columnwidth,height=\textheight,keepaspectratio]{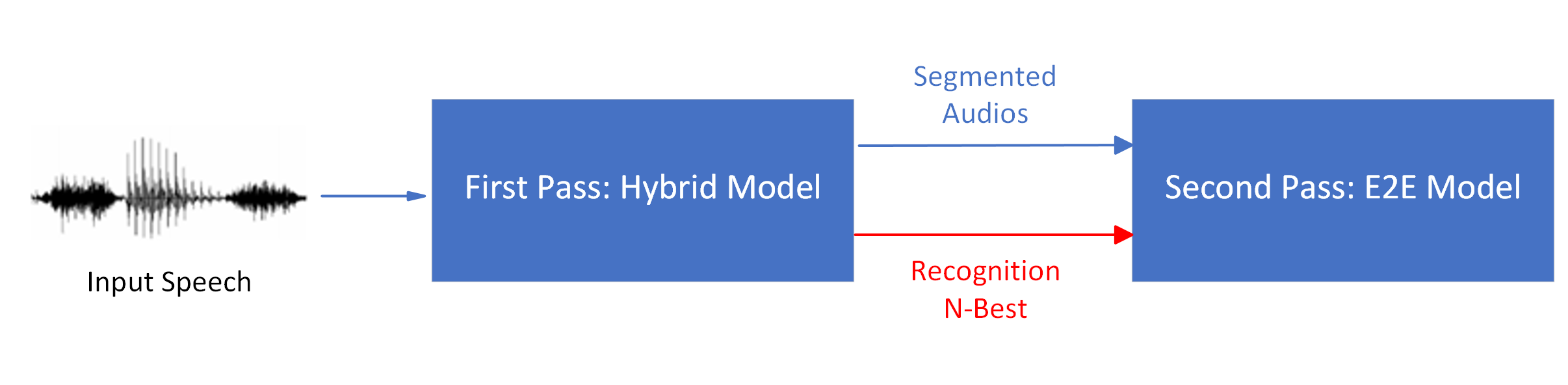}
    \caption{Two-pass \emph{Hybrid and E2E Cascading (HEC)} framework.}
    \label{fig:2_pass_structure}
\end{figure}

Unlike the previous research that combines two independently trained systems in parallel, in this paper, we propose a two-pass \emph{hybrid and E2E cascading (HEC)} framework as shown in Fig.~\ref{fig:2_pass_structure}. A conventional hybrid system is built as the first pass model, which outputs segmented audio and its corresponding N-best hypotheses. An attention-based encoder decoder (AED) model \cite{chorowski2014end, chan2016listen} is then trained as the second pass model that takes the segmented audio and N-best hypotheses as input and integrates them into the decoder module through attention mechanisms. By doing so, the two systems are combined more tightly as the second pass is totally aware of the first pass' output during training. This cascade design also creates opportunities for the second system to act like an error corrector of the first system which parallel combination systems can not do.

The work in this paper is inspired and shares many similarities with the deliberation model proposed in~\cite{hu2020deliberation, hu2021transformer}, which uses an RNN transducer (RNN-T) model \cite{he2019streaming} as first pass, whose output is fed into a second pass AED model. However, there are 2 key differences that distinguish our work. First, the deliberation model work combines two E2E systems, while we are combining the hybrid and E2E systems. The first pass hybrid model can do things that E2E model is not good at, such as making better use of an external LM and a customized biased list, as well as sophisticated speech segmentation. Those components are embedded in the output of the hybrid system, and passed to the second pass AED model in our work; while all the aforementioned challenges for E2E models still exist in the deliberation work which has E2E models in both passes. Also, both neural transducer and AED models in the deliberation work are in the same E2E model family, while hybrid and E2E models are two more distinct systems and expected to produce more complimentary error patterns. As shown in~\cite{wong2020combination}, combining hybrid and E2E models usually gives more gains than combining two E2E models.
Second, the encoder component of the two-pass models is shared in deliberation work, while there is no sharing in our work. Each design choice has its own advantages. Encoder sharing reduces computation cost and makes the model more attractive to latency/computation sensitive scenarios like on-device application. On the other hand, without sharing the encoder, it decouples the two-pass systems to some extent. As will be shown in Section~\ref{ssec:robust}, we can change the first pass model without retraining the second pass model, and still get reasonably good performance. This is particularly useful for real-world applications where the first pass hybrid model gets updated, while the second pass model does not have time to get updated. It is also a must-have feature for certain applications where the first pass model is a black-box API whose structure is unknown.  Furthermore, sharing the encoder may also introduce more similar error patterns of models in the two passes, thus it will be harder to achieve the best combination accuracy.

\vspace{-0.2cm}
\section{Two-Pass HEC Framework}
\vspace{-0.2cm}
\label{sec:BaseModel}
\subsection{Model Structure}
\subsubsection{First Pass Hybrid Model}
The acoustic model (AM) in the first pass system is the conventional tied-state Long Short-Term-Memory (LSTM) based hybrid model. An N-gram LM trained with a large amount of text-only data is used for decoding. A model-based speech segmentation module is also used to detect the end of speech during decoding.
 
\subsubsection{Second Pass AED Model}
\label{sssec:2ndpass_aed}
\begin{figure}[t]
    \centering
    \includegraphics[width=\columnwidth,height=\textheight,keepaspectratio]{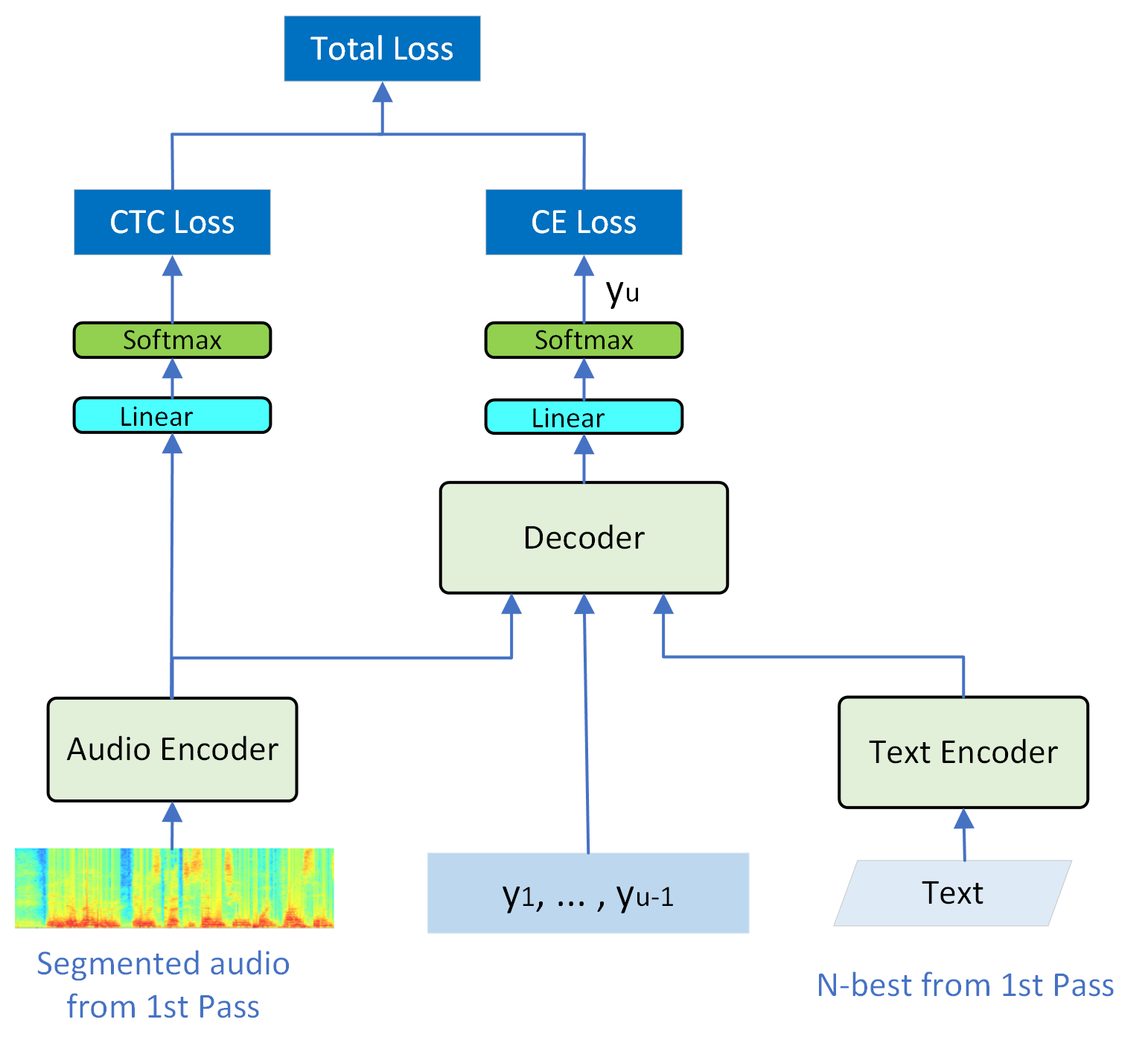}
    \caption{Second pass AED model in two-pass HEC Framework.}
    \label{fig:2nd_pass_aed}
    \vspace{-0.5cm}
\end{figure}
The second pass AED model consists of 3 components: audio encoder, text encoder, and decoder, as shown in Fig.~\ref{fig:2nd_pass_aed}. The audio encoder uses a multi-layer conformer encoder structure~\cite{gulati2020conformer} that converts the segmented audio from the first pass into a hidden vector sequence. Unlike the deliberation work in~\cite{hu2020deliberation}, this audio encoder does not share parameters with the first pass AM.

The text encoder adopts a multi-layer transformer encoder structure that converts the N-best hypotheses from the first pass into another hidden vector sequence. Although the N-best from the first-pass can be used, in practice, we find that the first best hypothesis is good enough. More N-best hypotheses are added in a similar way as~\cite{hu2021transformer}, but with only marginal gain found. To make things simple, in the rest of the paper, we only use one-best hypothesis as input to the text encoder.

The decoder predicts the $u$th position token output $y_u$ based on the previous predicted token sequence $y_1, ... , y_{u-1}$, as well as the two full hidden vector sequences generated by the audio and text encoders respectively.

\begin{figure}[t]
    \centering
    \includegraphics[width=\columnwidth,height=\textheight,keepaspectratio]{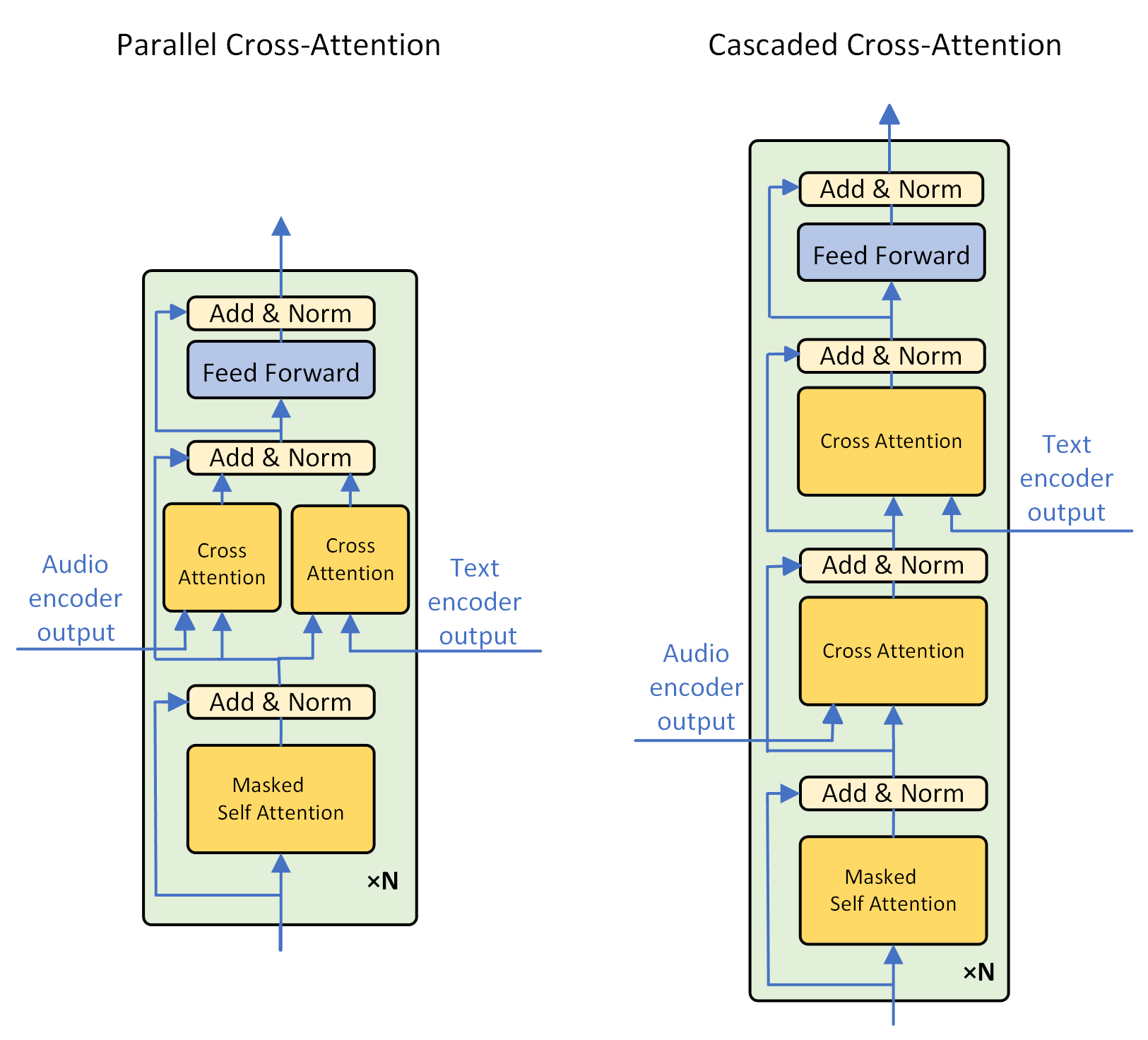}
    \caption{Two Structures to incorporate the audio and text encoders into the second-pass AED decoder: (left) Parallel  Cross Attention (PCA). (right) Cascaded Cross Attention (CCA).}
    \label{fig:2nd_pass_aed_decoder}
    \vspace{-0.5cm}
\end{figure}
The decoder follows the conventional transformer decoder structure, with self attention, cross attention and feedforward modules in each layer. To incorporate the information from the two encoder sources, we explore two different structures as shown in Fig.~\ref{fig:2nd_pass_aed_decoder}. In the ``Parallel Cross-Attention (PCA)" structure, two cross-attention components in each decoder layer attend to both outputs from audio and text encoders in parallel, and the context vectors obtained from both attention are linearly combined with equal weights. In the ``Cascaded Cross Attention  (CCA)" structure, cross attention is first applied to the audio encoder to generate a context vector. This context vector (after applying ``add \& norm") is then served as a query to do another cross attention to the text encoder. The comparison results between the two structures will be shown in Section~\ref{ssec:model_compare}.
\subsection{Training}
The training takes the following 3 steps.
\begin{itemize}
    \item STEP1: Given the audio and transcription pairs, the first pass hybrid AM is trained with a 3-stage model training recipe \cite{li2020high}: cross-entropy (CE), maximum mutual information (MMI)~\cite{vesely2013sequence}, as well as sequence-level teacher-student (T/S) learning \cite{wong2016sequence}. The LM and speech segmentation models are also prepared following the conventional recipe.
    \item STEP2: We decode the hybrid model trained in STEP1 on the whole training set. The one-best hypothesis for each training utterance are generated and stored.
    \item STEP3: The second-pass AED model is trained with audio and transcription pairs as well as the one-best hypothesis generated in STEP2 . During training, a joint connectionist temporal classification (CTC) \cite{graves2006connectionist} and attention CE loss is used as shown in Fig.~\ref{fig:2nd_pass_aed}. The combination weights are 0.3 and 0.7 for CTC and attention branches. The joint CTC and attention structure follows the work in~\cite{watanabe2017hybrid}, we find it works well especially for long utterances. This is another structure difference compared with the conventional AED model used in the second pass deliberation model, which was reported  to struggle with deletion errors decoding long utterances ~\cite{narayanan2021cascaded}. 
\end{itemize}

Unlike the deliberation model in \cite{hu2020deliberation}, the first pass and second pass models do not share parameters, and couple in a loose way. As a result, in the above training steps, the first pass and second pass models are allowed to have totally different structures, and  can be trained with different objective functions.

\subsection{Decoding}
 Given the input audio, the conventional hybrid system decoding is first carried out with the first pass AM, external LM (can be a customized LM), and the speech segmentation model. The output segmented speech audio and the corresponding N-best are then passed to the second pass AED model. The second pass model uses a one-pass beam search algorithm described in ~\cite{watanabe2017hybrid} to decode. During the beam search, the decoding score for each partial hypothesis is computed as a combination of CTC and attention scores, with the same combination weight as training.

\vspace{-0.2cm}
\section{Experiments}
\vspace{-0.2cm}
\label{sec:Experiments}
 
\subsection{Datasets}
We use 65 thousand (K) hours of transcribed Microsoft en-US data as the training data. The data is mainly from Microsoft production traffic, including Cortana and 
Conversational data which is a mixture of close-talk and far field speech recorded from a variety of devices.

Three test sets are used in the experiments. 
\begin{itemize}
    \item en-US General set: this test set covers 13 different application scenarios such as Cortana, far-field speech and call center from general American English speakers, consisting of a total of 1.8 million (M) words. This set is considered most matched to the training data.
    \item en-Dialect set: this set is also from Microsoft service traffic with a total of 1.2 M words, containing English speakers from the four countries: England, Australia, Canada and New Zealand.
    \item en-Accent set: this test set contains audio collected from English speakers with accent. It consists of 1.9 M words, and was recorded in well controlled clean environments.
\end{itemize}

All the training and test data is
anonymized  with personally identifiable information removed. The maximum duration of an utterance is 40 seconds.

We also do a full decoding of the whole 65K hour training data with the first pass hybrid model, and store the N-best hypotheses for training the second pass model.

\subsection{Modeling}
\vspace{0.2cm}

The first pass hybrid AM is an ensemble of two layer trajectory bi-directional LSTM
networks~\cite{sun2019layer}, with 6 layers of 1024 and 832 nodes, and 9404 senones as output. The model is trained in 3 stages, by CE, MMI and sequence-level T/S learning criterion respectively.
The input feature is 80-dimension log Mel filter bank for every 10 milliseconds (ms) speech. Frame skipping~\cite{miao2016simplifying} is applied by a factor of 2 to reduce the runtime cost. A 5-gram language model with around 1 million (M) vocabularies and 100 M n-grams is applied. This represents our best setup of offline (non-streaming) hybrid model. Notice that in many previous 2-pass work~\cite{gruenstein2021efficient,hu2020deliberation}, the first pass acoustic model is a streaming model, which represents the common application scenario. The reason that we use an offline model as first pass is to allow a fair comparison of the proposed method (whose second pass is offline) with the hybrid model, by excluding the accuracy gain from offline over streaming. Using offline AM as first-pass gives us a harder baseline to beat.

The second pass AED model consists of 3 components. The audio encoder consists of two convolutional layers that sub-sample the time
frame by a factor of 4, followed by 18 conformer layers. Each
conformer layer has a multi-head attention with 8 heads, and a
depth-wise convolution with kernel size of 3. The multi-head attention and the depth-wise convolution are sandwiched between
two 1024-dim feedforward layers. The text encoder consumes tokenized representation of the N-best with unigram sentencepice model~\cite{kudo-2018-subword} and consists of 6 conventional transformer encoder layers, with 2048-dim feedforward layer. The decoder consists of 6 layers , with 2048-dim feedforward layer as well. The embedding dimension is set to be 512 for all the 3 components. Two variants of models are built with different cross-attention structures: PCA and CCA as described earlier in Section~\ref{sssec:2ndpass_aed}.

For comparison, we also trained a conventional conformer-based AED model. Compared with the second pass AED model, the only difference is that there is no text encoder involved as there is no N-best input from the hybrid system. All the other structures and parameters are kept the same as the second pass AED model, with also a joint CTC-attention training objective function.

For both the second-pass and the conventional AED model, we use 80-dimensional log mel filter bank feature as input, without frame skipping (as the subsampling is done by the model itself).  4000 sentence pieces
are used as a recognition unit. No dropout is applied, which we found to give better model accuracy. Beam search with beam size 5 is used for model decoding.

\vspace{-0.3cm}
\section{Results}
\vspace{-0.2cm}
\label{sec:Results}

\subsection{Comparison of Different Models}
\vspace{0.2cm}

\label{ssec:model_compare}
\begin{table}
\centering
\setlength{\tabcolsep}{4.0pt}
\begin{tabular}[c]{|c||c|c|c|c|c|c|c|}
	\hline
	\multirow{2}{*}{\begin{tabular}{@{}c@{}} Test Set \end{tabular}}  
	& \multirow{2}{*}{\begin{tabular}{@{}c@{}} Hybrid \end{tabular}} 
	& \multirow{2}{*}{\begin{tabular}{@{}c@{}} AED \end{tabular}} 
	& \multicolumn{2}{c|}{HEC}
	& \multicolumn{2}{c|}{PCA WERR over}
	\\
	&  & & PCA & CCA & Hybrid & AED \\
	
    \hline
	
    en-US General & 8.37 & 7.27 & 6.86 & 6.84 & 18.0 & 5.6 \\
	\hline
    en-Dialect & 10.97 & 11.48 & 10.53 & 10.47 & 4.0 & 8.3 \\
	\hline
    en-Accent & 11.79 & 11.84 & 10.74 & 10.85 & 8.9 & 9.3 \\
	\hline
    Avg. & 10.31 & 10.04 & 9.24 & 9.26 & 10.4 & 8.0 \\
	\hline
	\end{tabular}
	\caption{WERs (\%) of Hybrid, AED, and two-pass HEC models with PCA or CCA decoder structure, on 3 test sets. WERR (\%) is the relative WER reduction of two-pass HEC-PCA model from Hybrid or AED models respectively. Avg. are word-count weighted average result over the 3 sets.}
\label{table:differet_model_wer}
\vspace{-15 pt}
\end{table}

\vspace{-0.2cm}

In Table~\ref{table:differet_model_wer}, we compare the conventional hybrid, conventional AED model, and the proposed two-pass HEC models (with PCA and CCA decoder structures) on the 3 test sets. For fair comparison, all models are non-streaming models, and trained with the same 65k hour data. 

By comparing the stand-alone hybrid and AED models, we observe that each model has its strength. The AED model performs significantly better than hybrid in the en-US general set. This is because the set matches well with the training data, and it is well known that E2E models demonstrate high accuracy for matching scenarios. On en-Dialect set, hybrid model starts to perform better as the training data mainly contains English speakers in North America, and does not match well with this dialect test set with non-American English speakers. The story on en-Accent set is more tricky: on one hand, the training data  does not contain a lot of en-Accent data, and hybrid model is expected to be better. On the other hand, the en-Accent test set is an engineered set with controlled recording condition which E2E model usually has good performance. As a result, the performance of hybrid and AED is almost on-par for this set. 

On every test set, the proposed two-pass HEC model performs better than each stand-alone hybrid and AED model. Two cross-attention structures CCA and PCA perform similarly. The overall WERR of HEC-PCA over hybrid and AED model are 10.4\% and 8.0\% respectively. This clearly demonstrates the combination benefits of the HEC structure.
\vspace{-0.3cm}
\subsection{Robustness to the First Pass Change}
\label{ssec:robust}
 We examine the robustness of the second pass model over the change of the first pass. There can be scenarios of having the first pass model updated either due to availability of new data or upgrade in the structure. This represents an unmatched scenario in testing, because the N-best is now generated from a new first pass model that has not been seen during second pass model training. Thus, we need to understand if it is safe to deploy the new first pass model, without seeing unexpected behavior in the second pass output. We empirically demonstrate that such update does not require retraining of the second pass model. 
 
 We trained a new hybrid model by adding more accent and dialect related data. The second-pass model was trained with the old hybrid model as the first pass. During testing, we use either old hybrid or new hybrid as the first pass. As shown in Table~\ref{table:second_pass_robust}, compared with the old hybrid model, due to adding matched training data,  the new hybrid model performs better on en-Dialect and en-Accent sets, with little degradation on general set. The two-pass HEC-PCA model still shows proportional gain by using the new hybrid model as the first pass, without retraining the second pass model.  Additional gain is not expected in this case, because the second pass model is not trained by adding the new data, also it is under a training/test mismatch case, i.e. it is  trained with N-best from the old hybrid, but test with N-best from the new hybrid. This shows high degree of robustness of the second pass model with respect to the first pass. This indicates that, in practice, we can deploy the updated first-pass model in a timely manner, without dependency on updating the second pass model. 

 \begin{table}
\centering
\setlength{\tabcolsep}{5.3pt}
\begin{tabular}[c]{|c||c|c|c|c|}
	\hline
	
	\multirow{2}{*}{\begin{tabular}{@{}c@{}} Test Set \end{tabular}}
	& \multirow{2}{*}{\begin{tabular}{@{}c@{}} Old \\ Hybrid \end{tabular}}
	& \multirow{2}{*}{\begin{tabular}{@{}c@{}} New \\ Hybrid \end{tabular}}
	& \multicolumn{2}{c|}{HEC-PCA}
 \\
	& & & Old Hyb & New Hyb \\
    \hline
    en-US General & 8.37 & 8.5 & 6.86 & 6.92 \\
	\hline
    en-Dialect & 10.97 & 10.47 & 10.53 & 10.21 \\
	\hline
    en-Accent & 11.79 & 10.06 & 10.74 & 10.11 \\
	\hline
    Avg. & 10.31 & 9.58 & 9.24 & 8.94 \\
	\hline
	\end{tabular}
 	\caption{WERs (\%) of two-pass HEC-PCA model with old and new first pass models. The second pass model is trained with old hybrid model as first pass, and test with old and new hybrid (Hyb) model as first pass.}
\label{table:second_pass_robust}
\vspace{-10 pt}
\end{table}


\subsection{Other Advantages of HEC over Pure E2E}
\label{ssec:other_advantage}
Besides the general recognition accuracy advantage, in this section, we will show some evidence of two specific advantages of the  HEC system over pure E2E model: speech segmentation and LM customization.
\subsubsection{The Speech Segmentation Effect on the Accuracy of Pure E2E }
\label{sssec:eos}
The test audio files in en-US General and en-Dialect sets are mostly from Microsoft traffic that are pre-segmented by hybrid system.  On these two test sets, we test our in-house streaming transformer transducer (T-T) model~\cite{chen2021developing} trained on the same 65k data by either depending on the end-of-speech (EOS) token to do segmentation and recognition, or just forcing it to decode through the whole  pre-segmented audio by hybrid system without using EOS. The reason we use this streaming T-T model rather than the previous offline AED model is that the EOS technique is more mature in this type of systems.  As shown in Table~\ref{table:eos}, introducing EOS detection in the E2E model in general hurts the speech recognition accuracy. For hybrid model, due to sophisticated development and optimization over the years, the situation can be less serious, thus benefiting HEC who directly uses segmented speech from first pass hybrid. Note that the WER results in Table~\ref{table:eos} should not be directly compared with previous Tables, as the T-T here is a streaming model with transformer encoder, while the previous AED models are offline conformer models.

 \begin{table}
\centering
\setlength{\tabcolsep}{4.0pt}
\begin{tabular}[c]{|c||c|c|c|c|c}
	\hline
	\multirow{2}{*}{\begin{tabular}{@{}c@{}} Test Set \end{tabular}}  
	& \multirow{2}{*}{\begin{tabular}{@{}c@{}} T-T \\ w/ EOS \end{tabular}} 
	& \multirow{2}{*}{\begin{tabular}{@{}c@{}} T-T \\ w/o EOS \end{tabular}} 
	& \multirow{2}{*}{\begin{tabular}{@{}c@{}} WERR \end{tabular}} 
   \\
	& & & \\
	\hline
    en-US General & 9.1 & 7.95 & 12.6 \\
	\hline
    en-Dialect & 16.94 & 15.54 & 8.3 \\
	\hline
	\end{tabular}
	\caption{WERs (\%) of streaming T-T models w/ and w/o EOS. WERR (\%) is the relative WER reduction of w/o EOS from w/ EOS. 
}
\label{table:eos}
\vspace{-15 pt}
\end{table}

\subsubsection{The Effect of LM customization in First Pass Hybrid}
We tested the two-pass HEC solution on a Microsoft internal task, where the first pass hybrid LM is heavily biased with the speaker-relevant name and entity list. Without doing anything to explicitly incorporate the name or entity information into second pass, we find that overall the second pass result does not have significant degradation on name and entity recognition, while improving significantly in general recognition accuracy with about 10\% WERR~\footnote{The content (audio, transcription and scorer) of the task is by policy requirement completely blind to us, thus we are not at liberty to disclose further experiment details and numbers.}. This indicates the system is indeed able to take advantage of the strength  of LM customization in hybrid modeling, which is a very challenging task for a pure E2E system.
\vspace{-0.2cm}
\section{Conclusion}
\vspace{-0.2cm}
\label{sec:Conclusion}
In this paper, we propose a two-pass HEC (hybrid and E2E cascading) framework to combine a hybrid and an E2E model. The proposed system is shown to give 8-10\% relative lower WER than each of the individual systems. Furthermore, the key advantages of each individual system can be largely kept, i.e., the language model customization with text-only data and sentence segmentation property of hybrid model, and the joint optimization property of the E2E model. We also show that the second pass model is robust with respect to the change of the first pass model, which is a nice property for real-time product deployment.
\vspace{-0.2cm}
\section{Acknowledgements}
\vspace{-0.2cm}
\label{sec:thanks}
The authors would like to thank Jeremy Wong and Eric Sun for providing the hybrid baseline, Yu Wu for providing the baseline T-T EOS model, Sourish Chatterjee and Amit Agarwal for internal set evaluation,  Akash Mahajan and Ruizhi Li for valuable comments, Tianyang Sun and Lingling Zhang for assisting with experiments.
\vfill\pagebreak

\bibliographystyle{IEEEbib}
\fontsize{8.7}{10.1}\selectfont
\bibliography{strings,refs}

\begin{thebibliography}{10}

\bibitem{soltau2016neural}
Hagen Soltau, Hank Liao, and Hasim Sak,
\newblock ``Neural speech recognizer: Acoustic-to-word lstm model for large
  vocabulary speech recognition,''
\newblock {\em arXiv preprint arXiv:1610.09975}, 2016.

\bibitem{li2018advancing}
Jinyu Li, Guoli Ye, Amit Das, Rui Zhao, and Yifan Gong,
\newblock ``Advancing acoustic-to-word ctc model,''
\newblock in {\em Proc. ICASSP}. IEEE, 2018, pp. 5794--5798.

\bibitem{chiu2018state}
Chung-Cheng Chiu, Tara~N Sainath, Yonghui Wu, et~al.,
\newblock ``State-of-the-art speech recognition with sequence-to-sequence
  models,''
\newblock in {\em Proc. ICASSP}. IEEE, 2018, pp. 4774--4778.

\bibitem{he2019streaming}
Yanzhang He, Tara~N Sainath, Rohit Prabhavalkar, et~al.,
\newblock ``Streaming end-to-end speech recognition for mobile devices,''
\newblock in {\em Proc. ICASSP}. IEEE, 2019, pp. 6381--6385.

\bibitem{zhang2020transformer}
Qian Zhang, Han Lu, Hasim Sak, Anshuman Tripathi, Erik McDermott, Stephen Koo,
  and Shankar Kumar,
\newblock ``Transformer transducer: A streamable speech recognition model with
  transformer encoders and rnn-t loss,''
\newblock in {\em Proc. ICASSP}. IEEE, 2020, pp. 7829--7833.

\bibitem{li2020comparison}
Jinyu Li, Yu~Wu, Yashesh Gaur, Chengyi Wang, Rui Zhao, and Shujie Liu,
\newblock ``On the comparison of popular end-to-end models for large scale
  speech recognition,''
\newblock {\em arXiv preprint arXiv:2005.14327}, 2020.

\bibitem{chen2021developing}
Xie Chen, Yu~Wu, Zhenghao Wang, Shujie Liu, and Jinyu Li,
\newblock ``Developing real-time streaming transformer transducer for speech
  recognition on large-scale dataset,''
\newblock in {\em Proc. ICASSP}. IEEE, 2021, pp. 5904--5908.

\bibitem{gruenstein2021efficient}
Tara~N. Sainath, Yanzhang He, Arun Narayanan, et~al.,
\newblock ``An efficient streaming non-recurrent on-device end-to-end model
  with improvements to rare-word modeling,''
\newblock 2021.

\bibitem{li2021recent}
Jinyu Li,
\newblock ``Recent advances in end-to-end automatic speech recognition,''
\newblock {\em arXiv preprint arXiv:2111.01690}, 2021.

\bibitem{toshniwal2018comparison}
Shubham Toshniwal, Anjuli Kannan, Chung-Cheng Chiu, Yonghui Wu, Tara~N Sainath,
  and Karen Livescu,
\newblock ``A comparison of techniques for language model integration in
  encoder-decoder speech recognition,''
\newblock in {\em Proc. SLT}. IEEE, 2018, pp. 369--375.

\bibitem{mcdermott2019density}
Erik McDermott, Hasim Sak, and Ehsan Variani,
\newblock ``A density ratio approach to language model fusion in end-to-end
  automatic speech recognition,''
\newblock in {\em Proc. ASRU}. IEEE, 2019, pp. 434--441.

\bibitem{meng2021internal}
Zhong Meng, Sarangarajan Parthasarathy, Eric Sun, Yashesh Gaur, Naoyuki Kanda,
  Liang Lu, Xie Chen, Rui Zhao, Jinyu Li, and Yifan Gong,
\newblock ``Internal language model estimation for domain-adaptive end-to-end
  speech recognition,''
\newblock in {\em Proc. SLT}. IEEE, 2021, pp. 243--250.

\bibitem{pundak2018deep}
Golan Pundak, Tara~N Sainath, Rohit Prabhavalkar, Anjuli Kannan, and Ding Zhao,
\newblock ``Deep context: end-to-end contextual speech recognition,''
\newblock in {\em Proc. SLT}. IEEE, 2018, pp. 418--425.

\bibitem{zhao2019shallow}
Ding Zhao, Tara~N Sainath, David Rybach, Pat Rondon, Deepti Bhatia, Bo~Li, and
  Ruoming Pang,
\newblock ``Shallow-fusion end-to-end contextual biasing.,''
\newblock in {\em Proc. Interspeech}, 2019, pp. 1418--1422.

\bibitem{wang2021light}
Xiaoqiang Wang, Yanqing Liu, Sheng Zhao, and Jinyu Li,
\newblock ``A light-weight contextual spelling correction model for customizing
  transducer-based speech recognition systems,''
\newblock {\em arXiv preprint arXiv:2108.07493}, 2021.

\bibitem{li2021better}
Bo~Li, Anmol Gulati, Jiahui Yu, et~al.,
\newblock ``A better and faster end-to-end model for streaming asr,''
\newblock in {\em Proc. ICASSP}. IEEE, 2021, pp. 5634--5638.

\bibitem{kim2021reducing}
Jaeyoung Kim, Han Lu, Anshuman Tripathi, Qian Zhang, and Hasim Sak,
\newblock ``Reducing streaming asr model delay with self alignment,''
\newblock {\em arXiv preprint arXiv:2105.05005}, 2021.

\bibitem{li2019integrating}
Qiujia Li, Chao Zhang, and Philip~C Woodland,
\newblock ``Integrating source-channel and attention-based sequence-to-sequence
  models for speech recognition,''
\newblock in {\em Proc. ASRU}. IEEE, 2019, pp. 39--46.

\bibitem{li2021combining}
Qiujia Li, Chao Zhang, and Philip~C Woodland,
\newblock ``Combining frame-synchronous and label-synchronous systems for
  speech recognition,''
\newblock {\em arXiv preprint arXiv:2107.00764}, 2021.

\bibitem{wong2020combination}
Jeremy~HM Wong, Yashesh Gaur, Rui Zhao, Liang Lu, Eric Sun, Jinyu Li, and Yifan
  Gong,
\newblock ``Combination of end-to-end and hybrid models for speech
  recognition.,''
\newblock in {\em Proc. Interspeech}, 2020, pp. 1783--1787.

\bibitem{chorowski2014end}
Jan Chorowski, Dzmitry Bahdanau, Kyunghyun Cho, and Yoshua Bengio,
\newblock ``End-to-end continuous speech recognition using attention-based
  recurrent {NN}: First results,''
\newblock {\em arXiv preprint arXiv:1412.1602}, 2014.

\bibitem{chan2016listen}
William Chan, Navdeep Jaitly, Quoc Le, and Oriol Vinyals,
\newblock ``Listen, attend and spell: A neural network for large vocabulary
  conversational speech recognition,''
\newblock in {\em Proc. ICASSP}. IEEE, 2016, pp. 4960--4964.

\bibitem{hu2020deliberation}
Ke~Hu, Tara~N Sainath, Ruoming Pang, and Rohit Prabhavalkar,
\newblock ``Deliberation model based two-pass end-to-end speech recognition,''
\newblock in {\em Proc. ICASSP}. IEEE, 2020, pp. 7799--7803.

\bibitem{hu2021transformer}
Ke~Hu, Ruoming Pang, Tara~N Sainath, and Trevor Strohman,
\newblock ``Transformer based deliberation for two-pass speech recognition,''
\newblock in {\em Proc. SLT}. IEEE, 2021, pp. 68--74.

\bibitem{gulati2020conformer}
Anmol Gulati, James Qin, Chung-Cheng Chiu, et~al.,
\newblock ``Conformer: Convolution-augmented transformer for speech
  recognition,''
\newblock {\em arXiv preprint arXiv:2005.08100}, 2020.

\bibitem{li2020high}
Jinyu Li, Rui Zhao, Eric Sun, Jeremy~HM Wong, Amit Das, Zhong Meng, and Yifan
  Gong,
\newblock ``High-accuracy and low-latency speech recognition with two-head
  contextual layer trajectory {LSTM} model,''
\newblock in {\em Proc. ICASSP}. IEEE, 2020, pp. 7699--7703.

\bibitem{vesely2013sequence}
Karel Vesel{\`y}, Arnab Ghoshal, Luk{\'a}s Burget, and Daniel Povey,
\newblock ``Sequence-discriminative training of deep neural networks.,''
\newblock in {\em Proc. Interspeech}, 2013, vol. 2013, pp. 2345--2349.

\bibitem{wong2016sequence}
JHM Wong and MJF Gales,
\newblock ``Sequence student-teacher training of deep neural networks,''
\newblock in {\em Proc. Interspeech}, 2016, pp. 2761--2765.

\bibitem{graves2006connectionist}
Alex Graves, Santiago Fern{\'a}ndez, Faustino Gomez, and J{\"u}rgen
  Schmidhuber,
\newblock ``Connectionist temporal classification: labelling unsegmented
  sequence data with recurrent neural networks,''
\newblock in {\em Proc. ICLM}. ACM, 2006, pp. 369--376.

\bibitem{watanabe2017hybrid}
Shinji Watanabe, Takaaki Hori, Suyoun Kim, John~R Hershey, and Tomoki Hayashi,
\newblock ``Hybrid ctc/attention architecture for end-to-end speech
  recognition,''
\newblock {\em IEEE Journal of Selected Topics in Signal Processing}, vol. 11,
  no. 8, pp. 1240--1253, 2017.

\bibitem{narayanan2021cascaded}
Arun Narayanan, Tara~N Sainath, Ruoming Pang, Jiahui Yu, Chung-Cheng Chiu,
  Rohit Prabhavalkar, Ehsan Variani, and Trevor Strohman,
\newblock ``Cascaded encoders for unifying streaming and non-streaming asr,''
\newblock in {\em Proc. ICASSP}. IEEE, 2021, pp. 5629--5633.

\bibitem{sun2019layer}
Eric Sun, Jinyu Li, and Yifan Gong,
\newblock ``Layer trajectory blstm.,''
\newblock in {\em Interspeech}, 2019, pp. 1403--1407.

\bibitem{miao2016simplifying}
Yajie Miao, Jinyu Li, Yongqiang Wang, Shi-Xiong Zhang, and Yifan Gong,
\newblock ``Simplifying long short-term memory acoustic models for fast
  training and decoding,''
\newblock in {\em Proc. ICASSP}. IEEE, 2016, pp. 2284--2288.

\bibitem{kudo-2018-subword}
Taku Kudo,
\newblock ``Subword regularization: Improving neural network translation models
  with multiple subword candidates,''
\newblock in {\em Proc. ACL}, Melbourne, Australia, July 2018, pp. 66--75,
  Association for Computational Linguistics.

\end{thebibliography}

\end{document}